# Making Sensitivity Analysis Computationally Efficient

Uffe Kjærulff
Department of Computer Science
Aalborg University
Fredrik Bajers Vej 7, DK-9220 Aalborg
Denmark
uk@cs.auc.dk

Linda C. van der Gaag
Department of Computer Science
Utrecht University
P.O. Box 80.089, 3508 TB Utrecht
The Netherlands
linda@cs.uu.nl

## Abstract

To investigate the robustness of the output probabilities of a Bayesian network, a sensitivity analysis can be performed. A one-way sensitivity analysis establishes, for each of the probability parameters of a network, a function expressing a posterior marginal probability of interest in terms of the parameter. Current methods for computing the coefficients in such a function rely on a large number of network evaluations. In this paper, we present a method that requires just a single outward propagation in a junction tree for establishing the coefficients in the functions for all possible parameters; in addition, an inward propagation is required for processing evidence. Conversely, the method requires a single outward propagation for computing the coefficients in the functions expressing all possible posterior marginals in terms of a single parameter. We extend these results to an $n$-way sensitivity analysis in which sets of parameters are studied.

## 1 INTRODUCTION

The robustness of the output probabilities of a Bayesian network can be investigated by performing a *sensitivity analysis* of the network. For mathematical models in general, sensitivity analysis serves to identify the effects of the inaccuracies in a model's parameters on its output (Morgan & Henrion 1990). For a Bayesian network, more specifically, performing a sensitivity analysis yields insight in the relation between the probability parameters of the network and its posterior marginals. The simplest type of sensitivity analysis is a one-way analysis in which a single parameter is studied; a more general $n$-way analysis serves to investigate the joint effects of inaccuracies in a set of parameters.

In the brute-force approach to performing a one-way sensitivity analysis of a Bayesian network, each probability parameter of the network is varied systematically and the effect on the output probabilities of the network is investigated. Performing a sensitivity analysis in this way requires thousands of network evaluations, or (full) propagations, and is, therefore, much too time consuming to be of any practical use.

Laskey (1995) has been the first to address the computational complexity of sensitivity analysis of Bayesian networks. She has introduced a method for computing the partial derivative of a posterior marginal probability with respect to a parameter under study. Her method thus yields a first-order approximation of the effect of varying a single probability parameter on a posterior marginal. Compared to the brute-force approach, her method requires considerably less computational effort. The method, however, provides insight only in the effect of small variations of parameters; when larger variations are considered, the quality of the approximation may rapidly break down.

The relation between a posterior marginal probability of interest and a parameter under study can be expressed through a simple mathematical function. The function expressing the posterior marginal is a quotient of two linear functions in the parameter, as has been shown by Castillo et al. (1996). Building upon this property, it suffices to establish the coefficients in this function to determine the effect of parameter variation. Castillo et al. (1997) and Coupé & van der Gaag (1998) have designed methods to this end. These methods require a single network evaluation for each coefficient to be established. Although these methods currently are the most efficient available, they rely on a large number of network evaluations and, as a consequence, are infeasible for large realistic networks.

In this paper, we present a new method for sensitivity analysis of Bayesian networks. Our method, like the two methods mentioned above, exploits the property that a posterior marginal probability relates by a simple mathematical function to a parameter under study. It requires just a single outward propagation in a junction tree, however, to compute the coefficients in the functions for all possible parameters; in addition, it requires an inward propagation for processing evidence. Conversely, the method requires a single outward propagation for establishing the coefficients in the functions expressing all possible posterior marginals in terms of a single parameter. Our method can be readily extended to an $n$-way sensitivity analysis in which sets of parameters are varied.

In addition to a sensitivity analysis, an *uncertainty analysis* can be performed for investigating the robustness of the output probabilities of a Bayesian network. In an uncertainty analysis, all parameters are varied simultaneously through sampling; it therefore provides little insight into the effects of variation of specific parameters. Experiments with uncertainty analysis have led to the suggestion that Bayesian networks are insensitive to inaccuracies in their parameters (Pradhan et al. 1996, Henrion et al. 1996). In these experiments, however, a measure of model robustness was obtained by assuming a lognormal distribution for each parameter and averaging over the probability of the true diagnosis for various diagnostic situations in a medical application. Rather than in the average of the probabilities of the true diagnosis, however, it is in the variation of these probabilities that inaccuracies in parameters are reflected. From these experimental results, therefore, no decisive conclusions can be drawn as to the sensitivity of Bayesian networks. In fact, Coupé et al. (1999) have reported high sensitivities in an emprical study in the medical domain, involving real patient data. We feel that these and emerging similar experiences warrant further investigation into sensitivity analysis of Bayesian networks.

The paper is organised as follows. Section 2 reviews the important basic property that a posterior marginal probability can be expressed as a quotient of two linear functions in a parameter under study. In Section 3, we briefly describe currently available methods for sensitivity analysis that build upon this property. In Section 4, we present our method for computing the coefficients in the functions for all possible parameters, using just one propagation in a junction tree. In Section 5, we describe a similar method for computing the coefficients in the sensitivity functions relating all possible posterior marginals to a single probability parameter. These methods are generalised to an $n$-way sensitivity analysis in Section 6. The paper ends with some concluding remarks in Section 7.

## 2 THE BASIC PROPERTY

Sensitivity analysis of a Bayesian network basically amounts to establishing, for each of the network's parameters, a function expressing an output probability in terms of the parameter under study. For output probabilities, we shall consider posterior marginal probabilities of the form $y = p(a|e)$, where $a$ is a value of a variable $A$ and $e$ denotes the evidence available. Each of the network's parameters is of the form $x = p(b_i | \pi)$, where $b_i$ is a value of a variable $B$ and $\pi$ is an arbitrary combination of values of the set of parents $\Pi = \text{pa}(B)$ of $B$. We will write $p(a|e)(x)$ to denote the function expressing the posterior marginal $p(a|e)$ in terms of the parameter $x$.

In the sequel, we will assume that in a sensitivity analysis, upon varying a parameter $x = p(b_i | \pi)$, each of the other probabilities $p(b_j | \pi)$ is co-varied accordingly, by scaling by the ratio between the probability masses left. More formally, let the variable $B$ have for its domain $\text{dom}(B) = \{b_1, \ldots, b_m\}$, $m \geq 1$. Note that the parameters $p(b_j | \pi)$, $j \neq i$, are functions of $x$. We now assume for these functions that

$$p(b_j | \pi)(x) = \begin{cases} x & \text{if } j = i \\ p(b_j | \pi) \dfrac{1-x}{1-p(b_i|\pi)} & \text{otherwise,} \end{cases} \quad (1)$$

with $p(b_i|\pi) < 1$.

With the assumption of co-variation as outlined above, the function $y(x)$ yielded by a sensitivity analysis is a quotient of two linear functions in $x$. The following theorem reviews this important property; the associated proof provides the basis for the algorithms presented in Sections 4 and 5.

**Theorem 1** *Let $p$ be the probability function defined by a Bayesian network over a set of variables $V$. Let $y = p(a|e)$ and $x = p(b_i|\pi)$ be as indicated above. Then,*

$$y = \frac{p(a,e)(x)}{p(e)(x)} = \frac{\alpha x + \beta}{\gamma x + \delta}, \quad (2)$$

*where $\alpha$, $\beta$, $\gamma$, and $\delta$ are constants with respect to $x$.*

*Proof:* The joint probability $p(a,e)$ can be expressed in terms of $x$ as

$$p(a,e)(x) = \left( \sum_{V:a,e} p(V) \right)(x),$$

where $\sum_{V:a,\ldots,d} p(V)$ denotes summation over the variables $V \setminus \{A, \ldots, D\}$ with $A, \ldots, D \in V$ fixed at values $a, \ldots, d$, respectively.

The sum $\sum_{V:a,e} p(V)$ in the above equation can be split into $n+1$ separate sums, such that the first sum includes only terms with the value $b_1$ for $B$ and the state $\pi$ for $\Pi$, the second sum includes only terms with the value $b_2$ for $B$ and $\Pi$ in state $\pi$, and so on, and the last sum includes the remaining terms. So,

$$p(a,e)(x)$$
$$= \left( \sum_j \sum_{V:a,e,b_j,\pi} p(V) + \sum_{V:a,e,\Pi \neq \pi} p(V) \right)(x)$$
$$= \sum_j p(b_j\,|\,\pi)(x) \frac{\sum_{V:a,e,b_j,\pi} p(V)}{p(b_j\,|\,\pi)} + \sum_{V:a,e,\Pi \neq \pi} p(V)$$
$$= x \frac{\sum_{V:a,e,b_i,\pi} p(V)}{p(b_i\,|\,\pi)} + \sum_{j \neq i} \left( \frac{p(b_j\,|\,\pi)}{1 - p(b_i\,|\,\pi)} - \right.$$
$$\left. x \frac{p(b_j\,|\,\pi)}{1 - p(b_i\,|\,\pi)} \right) \frac{\sum_{V:a,e,b_j,\pi} p(V)}{p(b_j\,|\,\pi)} + \sum_{V:a,e,\Pi \neq \pi} p(V)$$
$$= x \left( \frac{\sum_{V:a,e,b_i,\pi} p(V)}{p(b_i\,|\,\pi)} - \sum_{j \neq i} \frac{\sum_{V:a,e,b_j,\pi} p(V)}{1 - p(b_i\,|\,\pi)} \right) +$$
$$\sum_{j \neq i} \frac{\sum_{V:a,e,b_j,\pi} p(V)}{1 - p(b_i\,|\,\pi)} + \sum_{V:a,e,\Pi \neq \pi} p(V).$$

For the probability $p(e)$ we derive a similar expression by summing, in the above derivation, over all values of the variable $A$ instead of keeping it fixed at $a$. From the resulting expressions $p(a,e)(x)$ and $p(e)(x)$, it is readily seen that the output $y = p(a\,|\,e)$ can be written as a quotient of two functions that are linear in $x$. □

From Theorem 1 we have that the function that expresses a posterior marginal probability $y$ in terms of a single parameter $x$ is characterised by at most three coefficients. The theorem is easily extended to $n$ parameters. The function then includes the products of all possible combinations of parameters, termed monomials, in both its numerator and its denominator. The numerator as well as the denominator are characterised by $2^n$ coefficients, many of which may be zero.

## 3　CURRENT METHODS

The most efficient methods for sensitivity analysis of Bayesian networks currently available exploit the basic property reviewed in the previous section. We briefly review these methods.

Not all parameters in a Bayesian network can influence a posterior marginal probability of interest. The subset of parameters (possibly) influencing the posterior marginal is dependent upon the evidence $e$. The set of relevant parameters is easily identified using a variation of the algorithm described by Geiger et al. (1990), as described by Castillo et al. (1997). After having identified the set of relevant parameters, the sensitivity analysis can be restricted to this set.

Building upon the set of $n$ relevant parameters, $x_1, \ldots, x_n$, the algorithm of Castillo et al. (1997) identifies sets of monomials for which the coefficients will be zero in the linear function $p(a,e)(x_1, \ldots, x_n)$. For the resulting $m$ monomials, the algorithm constructs a system of $m$ independent equations of the form $y^i = p(a,e)(x_1^i, \ldots, x_n^i)$, where, for each $j$, $x_j^i$ denote arbitrary values for parameter $x_j$. The corresponding values $y^i$, $i = 1, \ldots, m$, are obtained through $m$ network evaluations. The coefficients in the function are now determined by solving the set of equations thus obtained. Coupé & van der Gaag independently described a similar method, also based on the idea of solving a system of independent equations. They further argue that in a one-way sensitivity analysis three network evaluations suffice per relevant parameter.

The methods reviewed above have a computational complexity that is considerably less than the brute-force approach of systematic variation of parameters. However, the methods can still be quite time consuming: for a network of realistic size, it can easily require several hundreds of network evaluations to perform a one-way sensitivity analysis. An more general $n$-way sensitivity analysis to study the joint effect of simultaneous variation of $n$ parameters can in fact be so time consuming that it is infeasible in practice.

## 4　ANALYSIS OF ONE OUTPUT WRT. ALL PARAMETERS

The new methods for sensitivity analysis presented in this paper have been tailored to Bayesian networks in their junction-tree representation. The methods basically perform a single or a few outward propagations in a junction tree and, as a result, are much less time consuming than the methods reviewed in the previous section.

In this section, we present our method for computing the coefficients in the sensitivity functions expressing a posterior marginal of interest $y = p(a\,|\,e)$ in terms of all possible probability parameters $x$. Recall that these functions are of the form presented in Theorem 1. Our method now builds on the idea that, in a junction tree, the expressions for $p(a,e)$ and $p(e)$ in terms of $x$ can be derived from the potential of a clique containing both the variable and the parents to which the parameter $x$ pertains. The following theorem details the coefficients to be computed.

**Theorem 2** *Let $p$ be the probability function defined by a Bayesian network and let $\mathcal{T}$ be a junction tree*

for the network. Let $y = p(a|e)$ and $x = p(b_i|\pi)$ be as before. Suppose that, in $\mathcal{T}$, an inward propagation has been performed towards a clique containing the variable of interest $A$; suppose that subsequently an outward propagation from this clique has been performed with the value $a$ for $A$. Now, let $K$ be a clique in $\mathcal{T}$ containing both the variable $B$ and its parents $\Pi = pa(B)$; let $\phi_K = p(K, a, e)$ be the potential of clique $K$ after the abovementioned propagations. Then, $p(a,e)(x) = \alpha x + \beta$ with

$$\alpha = \frac{\sum_{K:b_i,\pi} \phi_K}{p(b_i|\pi)} - \sum_{j \neq i} \frac{\sum_{K:b_j,\pi} \phi_K}{1 - p(b_i|\pi)}, \quad (3)$$

$$\beta = \sum_{j \neq i} \frac{\sum_{K:b_j,\pi} \phi_K}{1 - p(b_i|\pi)} + \sum_{K:\Pi \neq \pi} \phi_K. \quad (4)$$

Proof: The property follows directly from the proof of Theorem 1 by observing that $p(a,e) = \sum_K \phi_K$. □

Building upon similar observations, we have the following corollary.

**Corollary 1** Let $p$ be the probability function defined by a Bayesian netwerk and let $\mathcal{T}$ be a junction tree for the network. Let $x = p(b_i|\pi)$ and $K$ be as before. Suppose that the evidence $e$ has been processed in $\mathcal{T}$ by an inward and subsequent outward propagation. Let $\phi_K^* = p(K, e)$ be the potential of clique $K$ after the propagation. Then, $p(e)(x) = \gamma x + \delta$ with

$$\gamma = \frac{\sum_{K:b_i,\pi} \phi_K^*}{p(b_i|\pi)} - \sum_{j \neq i} \frac{\sum_{K:b_j,\pi} \phi_K^*}{1 - p(b_i|\pi)}, \quad (5)$$

$$\delta = \sum_{j \neq i} \frac{\sum_{K:b_j,\pi} \phi_K^*}{1 - p(b_i|\pi)} + \sum_{K:\Pi \neq \pi} \phi_K^*. \quad (6)$$

Theorem 2 and Corollary 1 provide the basis for our method for computing the coefficients in the functions expressing the posterior marginal of interest $y = p(a|e)$ in terms of all possible parameters $x$. The method is composed of the following steps:

1. Enter the evidence $e$ into the junction tree and perform an inward and an outward propagation using an arbitrary root clique.

2. Compute the coefficients $\gamma$ and $\delta$, using the equations (5) and (6) from Corollary 1, for all relevant parameters, locally per clique.

3. Perform an outward propagation from a clique containing the variable of interest $A$, with the additional evidence $A = a$.

4. Compute the coefficients $\alpha$ and $\beta$, using the equations (3) and (6) from Theorem 2, for all relevant parameters, locally per clique.

We would like to note that our method requires just one inward and two outward propagations to establish *all* sensitivity functions for a specific posterior marginal, whereas the methods reviewed in the previous section require three inward and outward propagations per parameter.

The method described above outlines the basic idea. The method, however, may be easier to implement in the alternative form based upon Theorems 3 and 4.

**Theorem 3** Let the junction tree $\mathcal{T}$ be as before. Also, let $y = p(a|e)$ and $x = p(b_i|\pi)$ be as before and let $K$ be a clique in $\mathcal{T}$ including both $B$ and $\Pi = pa(B)$. Now, let $x^1$ be the initially specified value for $x$ and let $x^2$ denote an arbitrary other value for $x$. Suppose that, in $\mathcal{T}$, an inward propagation has been performed towards a clique containing the variable of interest $A$; suppose that subsequently an outward propagation has been performed with the value $a$ for $A$. Now, let $\phi_K = p(K, a, e)$ be the resulting clique potential for clique $K$. Let

$$y^1 = p(a,e)(x^1) = \sum_K \phi_K, \quad (7)$$

$$y^2 = p(a,e)(x^2) = \sum_K \phi_K \frac{p'(B|\pi)}{p(B|\pi)}, \quad (8)$$

where $p(B|\pi)$ and $p'(B|\pi)$ denote parameter vectors with $x = x^1$ and $x = x^2$, respectively. Then, $y = \alpha x + \beta$ with

$$\alpha = \frac{y^1 - y^2}{x^1 - x^2} \quad \text{and} \quad \beta = \frac{x^1 y^2 - x^2 y^1}{x^1 - x^2}. \quad (9)$$

Proof: Since both variable $B$ and its parents are included in clique $K$, we can obtain from the parameter vector

$p(B|\pi)$
$= (q_1(x^1), \ldots, q_{i-1}(x^1), x^1, q_{i+1}(x^1), \ldots, q_n(x^1))$

the parameter vector

$p'(B|\pi)$
$= (q'_1(x^2), \ldots, q'_{i-1}(x^2), x^2, q'_{i+1}(x^2), \ldots, q'_n(x^2))$

where $q$ and $q'$ are parameters co-varying according to equation (1), by multiplication of the potential $\phi_K$ by $p'(B|\pi)/p(B|\pi)$. From Theorem 1, we have that $y = \alpha x + \beta$. The expressions for $\alpha$ and $\beta$ now follow from simple mathematical manipulation. □

**Theorem 4** Let the junction tree $\mathcal{T}$ be as before. Also, let $y = p(a|e)$ and $x = p(b_i|\pi)$ be as before. Suppose that, in $\mathcal{T}$, an inward propagation has been performed towards a clique including the variable of interest $A$. Then, $p(e)(x) = \gamma x + \delta$ with

$$\gamma = \alpha_a + \alpha_{\neg a} \quad \text{and} \quad \delta = \beta_a + \beta_{\neg a}, \qquad (10)$$

where $\alpha_a, \beta_a$ and $\alpha_{\neg a}, \beta_{\neg a}$ are as in equation (9), obtained from outward propagations with the evidence $A = a$ and $A \neq a$, respectively.

*Proof:* We begin by observing that $p(e)(x) = p(a,e)(x) + p(\neg a, e)(x)$. By entering the evidence $A = a$ in a clique $H$ containing the variable $A$ and propagating outwards, we obtain the potential $\phi_K = p(K, a, e)$ for clique $K$. From this potential, $y_a^1 = p(a, e) = \sum_K \phi_K$ is readily computed, as described in equations (7) from Theorem 3. Similarly, by entering the evidence that $A$ does not have the value $a$ (that is, by multiplying the clique potential for $H$ with a vector over $\text{dom}(A)$, in which the entry corresponding to state $a$ is zero and all other entries equal 1) and propagating outwards, we obtain the potential $\phi'_K = p(K, \neg a, e)$. From this potential, $y_{\neg a}^1 = p(\neg a, e) = \sum_K \phi'_K$ is readily computed. Using equation (8) from Theorem 3, we get $y_a^2$ and $y_{\neg a}^2$. Now, using equation (9), we find $\alpha_a$, $\alpha_{\neg a}$, $\beta_a$, and $\beta_{\neg a}$. Inserting these coefficients into the expression $p(e)(x) = p(a, e)(x) + p(\neg a, e)(x)$ yields the result stated in the theorem. □

Our alternative method, building upon Theorems 3 and 4, is composed of the following steps:

1. Enter the evidence $e$ into the junction tree and perform an inward propagation towards a clique $H$ containing the variable of interest $A$.

2. Perform an outward propagation from $H$ with the additional evidence $A = a$.

3. Compute $y_a^1$ and $y_a^2$, using the equations (7) and (8) from Theorem 3.

4. Compute the coefficients $\alpha \equiv \alpha_a$ and $\beta \equiv \beta_a$, using (9), for all relevant parameters, locally per clique.

5. Retract the evidence $A = a$ without retracting the evidence $e$.

6. Perform an outward propagation from $H$ with the additional evidence $A \neq a$.

7. Compute $y_{\neg a}^1$ and $y_{\neg a}^2$, using the equations (7) and (8) from Theorem 3.

8. Compute $\alpha_{\neg a}$ and $\beta_{\neg a}$, using (9).

9. Compute the coefficients $\gamma$ and $\delta$, using equation (10) from Theorem 4.

To allow for retracting the evidence $A = a$ in Step 5 of our method without retracting $e$, fast-retraction propagation (Cowell & Dawid 1992) is used in Step 2.

Comparing the computational costs of the two alternative methods, we note that they both require one inward and two outward propagations. Considering the first method, we observe that Steps 2 and 4 are equally costly. The computation of the coefficients $\alpha$ and $\gamma$ costs $2 \cdot |\text{dom}(K)|/m$ operations, where $m = |\text{dom}(\text{pa}(B))|$; the computation of $\beta$ and $\delta$ costs $2 |\text{dom}(K)|$ arithmetic operations. Thus, the additional cost of the first method is roughly in the order of 2 to 3 times $|\text{dom}(K)|$. The additional costly steps in the second method are Steps 3 and 7, both costing approximately $3 \cdot |\text{dom}(K)|$ operations. Thus, the additional cost of the second method is roughly $6 \cdot |\text{dom}(K)|$ arithmetic operations. This rough comparison of the computational costs of the two methods only addresses the number of arithmetic operations involved. The first method, however, has a much larger overhead in terms of computing indices in performing the various summations. Thus, depending on the implementation, the two methods might very well have comparable performance.

## 5 ANALYSIS OF ALL OUTPUTS WRT. ONE PARAMETER

Having identified a parameter $x$ to which an output probability of a Bayesian network is particularly sensitive, one might be interested in establishing sensitivity functions for all possible posterior marginals in terms of this parameter. Such an analysis amounts to computing the coefficients in these functions. Note that while the method described in Section 4 provides for evaluating the overall robustness of a Bayesian network, the method described in this section is provides for getting insight in the spread of influence from separate parameters.

The method of Castillo et al. (1997) and the method of Coupé & van der Gaag (1998) can be exploited for establishing the coefficients in the sensitivity functions for all possible output probabilities, requiring three propagations per posterior marginal. Building upon the ideas put forward in the previous section, however, a more efficient method is obtained. Theorem 5 provides the basis for our method.

**Theorem 5** Let the junction tree $\mathcal{T}$ be as before. Also, let $y = p(a|e)$ and $x = p(b|\pi)$ be as before. Suppose that, in $\mathcal{T}$, an inward propagation has been performed towards a clique containing the variable $B$ to which the parameter $x$ pertains. Then, $p(e)(x) = \gamma x + \delta$ with

$$\gamma = \alpha_a + \alpha_{\neg a} \text{ and } \delta = \beta_a + \beta_{\neg a}, \quad (11)$$

where $\alpha_a, \beta_a$ and $\alpha_{\neg a}, \beta_{\neg a}$ are as in equation (9), obtained from two outward propagations with two distinct values for $x$.

*Proof:* Let $K$ be a clique containing both the variable $B$ and its set of parents $\Pi = \text{pa}(B)$. Let $x^1$ and $x^2$ denote two different values of the parameter $x$. From the outward propagation using $x^1$ from clique $K$, we obtain the probability vector $p(A,e)(x^1) = (y_a^1, y_{\neg a}^1)$ through marginalization of the clique potential for a clique $H$ containing $A$. Similarly, from the outward propagation with $x^2$, we find the vector $p'(A,e)(x^2) = (y_a^2, y_{\neg a}^2)$. From

$$p(A,e)(x^1) = \begin{pmatrix} y_a^1 \\ y_{\neg a}^1 \end{pmatrix} = \begin{pmatrix} \alpha_a \\ \alpha_{\neg a} \end{pmatrix} x^1 + \begin{pmatrix} \beta_a \\ \beta_{\neg a} \end{pmatrix}$$

$$p'(A,e)(x^2) = \begin{pmatrix} y_a^2 \\ y_{\neg a}^2 \end{pmatrix} = \begin{pmatrix} \alpha_a \\ \alpha_{\neg a} \end{pmatrix} x^2 + \begin{pmatrix} \beta_a \\ \beta_{\neg a} \end{pmatrix}$$

we get the result stated in the theorem. □

Theorem 5 provides the basis for our method for computing the coefficients in the functions expressing all possible output probabilities $y = p(a|e)$ in a single parameter $x = p(b|\pi)$. The method is composed of the following steps:

1. Enter the evidence $e$ into the junction tree and perform an inward and an outward propagation using an arbitrary root clique.

2. Compute the probability vector $p(A,e) = (y_a^1, y_{\neg a}^1)$ through marginalization of the clique potential for a clique $H$ containing $A$, for all variables of interest.

3. Change the value of parameter $x$ and perform an outward propagation from a clique containing both the variable $B$ and its parents.

4. Compute the probability vector $p'(A,e) = (y_a^2, y_{\neg a}^2)$ through marginalization of the potential for clique $H$, for all variables of interest.

5. Compute $\alpha_a, \alpha_{\neg a}, \beta_a,$ and $\beta_{\neg a}$, using the equation (9) from Theorem 3.

6. Compute $\gamma$ and $\delta$, using the equation (11).

We would like to note that our method requires just one inward and two outward propagations to establish *all* sensitivity functions for a specific probability parameter.

## 6 $n$-WAY SENSITIVITY ANALYSIS

So far, we have addressed one-way sensitivity analyses only, in which the effects of separate parameters are studied. In this section, we turn our attention to more general $n$-way analyses in which the effects of simultaneous variation of $n$ parameters are studied.

One can regard $n$-way sensitivity analysis as involving analyses of joint effects for all subsets of size $n$ or less of all, say $m$, relevant parameters. Using the method of Castillo et al. (1997), this would involve $\sum_{i=1}^{n} \binom{m}{i}$ separate analyses and $\sum_{i=1}^{n} r^i \binom{m}{i}$ probability propagations to compute the $2^n$ coefficients, assuming $r$-ary variables.

Provided the $n$ parameters all belong to the same clique, Theorem 6 below states that we only need one propagation to compute the $2^n$ coefficients, but $\sum_{i=1}^{n} \binom{m}{i}$ local computations involving marginalizations of clique potentials.

In essence, Theorem 1 states that the mathematical expression for a probability, $p(e)(x)$, of a vector of instantiations, $e$, as a function of a probability parameter, $x$, takes the form of a linear function of $x$. Theorem 6 generalizes this statement to the case with $n$ parameters, $x_1, \ldots, x_n$, and states that the resulting function is a multilinear function in $x_1, \ldots, x_n$.

To simplify the exposition, we shall assume that the parameters are independent; that is, for each pair of parameters, $x_i = p(b_{x_i}|\pi_{x_i})$ and $x_j = p(b_{x_j}|\pi_{x_j})$, with the associated variables $B_{x_i}, \Pi_{x_i}, B_{x_j},$ and $\Pi_{x_j}$, it holds true that $\pi_{x_i} \neq \pi_{x_j}$, $B_{x_i} \notin \Pi_{x_j}$, and $B_{x_j} \notin \Pi_{x_i}$. To generalize the theorem to cover the case of dependent parameters is fairly straightforward.

**Theorem 6** *Let $p$ be the probability function for a Bayesian network, where evidence $e$ has been propagated in a junction tree, $\mathcal{T}$, for the network. Let $X = \{x_1, \ldots, x_n\}$ be a set of parameters of the network, where, for each $i = 1, \ldots, n$,*

$$x_i = p(b_{x_i}|\pi_{x_i}),$$

*with the associated variables, $B_{x_i}$ and $\Pi_{x_i}$, being members of a clique, $C$, in $\mathcal{T}$. Then*

$$p(e)(X) = \sum_{Z \subseteq X} \gamma_X(Z) \prod_{z \in Z} z + \sum_{C:\Pi \neq \pi} \phi_C,$$

*where $\Pi = \{\Pi_{x_1}, \ldots, \Pi_{x_n}\}$, $\pi = (\pi_{x_1}, \ldots, \pi_{x_n})$, and*

$$\gamma_X(Z) = (-1)^{|X \setminus Z|} \sum_{Y \subseteq Z} f_X(Y),$$

*where*

$$f_X(Y) = (-1)^{|X \setminus Y|}$$

$$\sum_{b'_{w_1} \neq b_{w_1}} \cdots \sum_{b'_{w_k} \neq b_{w_k}} \frac{\sum_{C:b_Y,b'_W,\pi} \phi_C}{\prod_{y \in Y} p_y \prod_{w \in W}(1-p_w)},$$

where $\phi_C = p(C,e)$, $W = X \setminus Y = \{w_1, \ldots, w_k\}$, $k = |W|$, $b_Y = (b_{y_1}, \ldots, b_{y_{|Y|}})$, $b'_W = (b'_{w_1}, \ldots, b'_{w_{|W|}})$, and $p_{x_i}$ denotes the initial value of $x_i$.

*Proof:* Using the same procedure as in the proof of Theorem 1 we get

$$p(e)(X)$$

$$= \left(\sum_C \phi_C\right)(X)$$

$$= \left(\sum_{b'_{x_1}} \cdots \sum_{b'_{x_n}} \sum_{C:b'_{x_1},\ldots,b'_{x_n},\pi} \phi_C + \sum_{C:\Pi \neq \pi} \phi_C\right)(X)$$

$$= \sum_{b'_{x_1}} \cdots \sum_{b'_{x_n}} p(b'_{x_1}|\pi_{x_1})(x_1) \cdots p(b'_{x_n}|\pi_{x_n})(x_n)$$

$$\frac{\sum_{C:b'_{x_1},\ldots,b'_{x_n},\pi} \phi_C}{p(b'_{x_1}|\pi_{x_1}) \cdots p(b'_{x_n}|\pi_{x_n})} + \sum_{C:\Pi \neq \pi} \phi_C.$$

The multiple sum can be grouped into sums over the subsets $Z \subseteq X$ such that $b'_z = b_z$ for each $z \in Z$:

$$p(e)(X) = \sum_{Z \subseteq X} \prod_{z \in Z} z \sum_{b'_{u_1} \neq b_{u_1}} \cdots \sum_{b'_{u_{|U|}} \neq b_{u_{|U|}}}$$

$$p(b'_{u_1}|\pi_{u_1})(u_1) \cdots p(b'_{u_{|U|}}|\pi_{u_{|U|}})(u_{|U|})$$

$$\frac{\sum_{C:b_Z,b'_U,\pi} \phi_C}{\prod_{z \in Z} p(b_z|\pi_z) \prod_{u \in U} p(b'_u|\pi_u)} +$$

$$\sum_{C:\Pi \neq \pi} \phi_C, \qquad (12)$$

where $U = X \setminus Z = \{u_1, \ldots, u_{|U|}\}$. Now, expanding the terms $p(b'_u|\pi_u)(u)$, $u \in U$, using (1), an easy calculation yields

$$p(b'_{u_1}|\pi_{u_1})(u_1) \cdots p(b'_{u_{|U|}}|\pi_{u_{|U|}})(u_{|U|})$$

$$= \prod_{u \in U} \left(\frac{p(b'_u|\pi_u)}{1-p(b_u|\pi_u)} - u \frac{p(b'_u|\pi_u)}{1-p(b_u|\pi_u)}\right)$$

$$= \prod_{u \in U} \frac{p(b'_u|\pi_u)}{1-p(b_u|\pi_u)} \left(\sum_{S \subseteq U} (-1)^{|S|} \prod_{x \in S} x\right). \quad (13)$$

Inserting (13) in (12) and rearranging terms yields

$$p(e)(X) = \sum_{Z \subseteq X} \prod_{z \in Z} z \sum_{S \subseteq U} (-1)^{|S|} \prod_{x \in S} x$$

$$\sum_{b'_{u_1} \neq b_{u_1}} \cdots \sum_{b'_{u_{|U|}} \neq b_{u_{|U|}}} \frac{\sum_{C:b'_U,b_Z,\pi} \phi_C}{\prod_{z \in Z} p_z \prod_{u \in U}(1-p_u)} +$$

$$\sum_{C:\Pi \neq \pi} \phi_C. \qquad (14)$$

If, instead of summing over subsets $S \subseteq X \setminus Z$, we sum over the subsets of $Z$ and takes care that the signs of the terms are preserved, we get the desired result. □

Note that, since the computation of $\gamma_X(X)$ ranges over all subsets of $X$, $\gamma_X(Z)$ can be computed, for each $Z \subset X$, as a sum over a subset of the terms involved in the computation of $\gamma_X(X)$.

The result presented in Theorem 6 is limited in the sense that all parameters under investigation must belong to the same clique in the junction tree. We shall now present a more general method which utilizes the results obtained by lower-order analyses. Let $X = \{x_1, \ldots, x_n\}$ be the $n$ parameters under investigation and write

$$p(e)(X) = \sum_{Z \subseteq X} \gamma(Z) \prod_{z \in Z} z + \delta. \qquad (15)$$

Through one-way analysis involving the parameter $x$ we obtain the constants $\alpha_x$ and $\beta_x$ in

$$p(e)(x) = \alpha_x x + \beta_x.$$

The $2^n$ constants in (15) are related to $\alpha_x$ and $\beta_x$ in the way that $\alpha_x$ equals the sum of all coefficients, $\gamma(Z)$, for which $x \in Z$, and $\beta_x$ equals the sum of the remaining coefficients. That is,

$$\alpha_x = \sum_{Z \subseteq X : x \in Z} \gamma(Z) \qquad (16)$$

and

$$\beta_x = \sum_{Z \subseteq X : x \notin Z} \gamma(Z) + \delta. \qquad (17)$$

Thus, for each one-way analysis of a parameter $x_i$, $i = 1, \ldots, n$, we obtain 2 equations of the form (16) and (17). In addition, we obtain the equation

$$p(e) = \sum_{Z \subseteq X} \gamma(Z) \prod_{z \in Z} z_0 + \delta, \qquad (18)$$

where $z_0$ denote the value specified for parameter $z$ in the Bayesian network. This gives us a total of $2n+1$ equations. However, we need (at least) $2^n$ equations to compute the $2^n$ coefficients.

Now, if for each parameter, $z \in Z$, we assign a new value and perform a full propagation, then we obtain an additional $2n+1$ equations. Thus, we can obtain at least $2^n$ equations by performing

$$\left\lfloor \frac{2^n}{2n+1} \right\rfloor$$

full propagations with different parameter values, in addition to the initial propagation performed for the one-way analyses. For example, to perform 4-way analyses, a total of two full propagations is sufficient.

This result can be generalized very easily, since each $m$-way analysis gives rise to $2^m$ equations of the form (16) and (17). Thus, to perform $n$-way analyses, where $n > m$, we need at most

$$\left\lfloor 2^{n-m} / \binom{n}{m} \right\rfloor$$

additional propagations, as there are $\binom{n}{m}$ relevant $m$-way analyses. So, for example, if we have performed 2-way analyses and want to perform 5-way analyses, no further propagations are needed.

## 7 CONCLUDING REMARKS

We have presented methods for sensitivity analysis of Bayesian networks which are significantly more efficient than current methods. In the case of one-way analysis, the number of probability propagations of current methods grows linearly in the number of relevant parameters, whereas the methods presented above only requires one inward and one or two outward propagations, no matter the number of relevant parameters.

To substantiate the importance of this difference, we have investigated three real-world networks to get an idea of the typical number of relevant parameters in a realistic scenario. All three networks are from the medical domain: a subnetwork of Munin (Andreassen et al. 1989) containing 1003 variables, a network modelling the pathophysiology of ventricular septal defect (Coupé et al. 1999) containing 38 variables, and a network related to disorders in the oesophagus containing 70 variables. The investigation were conducted using real patient data involving, respectively, 15, 5, and 3 patients. The average number of relevant parameters were found to be 16313, 738, and 992, respectively. (Since no censoring of parameters representing functional relationships were performed on parameters for the Munin network, the figure 16313 is probably somewhat overestimated.)

Efficient methods for sensitivity analysis play an important role in both the knowledge acquisition and the validation phases for manually constructed Bayesian-network models.

Coupé et al. (1999) reports on an empirical study using sensitivity analysis to focus attention on the most influential parameters in the knowledge acquisition process, thereby considerably reducing the time required to acquire the parameter values.

The validation phase involves two aspects: fine-tuning and robustness analysis. The fine-tuning aspect involves adjustment of the parameter values to make the network respond correctly to a number of test cases. A gradient descent approach is useful for that purpose (cf. neural network type training), where the gradient of a posterior marginal with respect to a subset of parameters can easily be computed through a minor modification of the algorithms for sensitivity analysis. Based on the work described in the present paper, Jensen (1999) has suggested a method for gradient descent training of Bayesian networks.

Once a network has been fine-tuned, and thus responds correctly on a selection of test cases, the robustness of the network may be investigated. This involves, in essence, determining lower and upper bounds for parameter values for which the output of the network still agrees with the test cases. A parameter value close to one of the bounds indicate a possible lack of robustness. Given analytic expressions for the outputs in terms of the parameters, derived by e.g. methods described in the present paper, these bounds are easily determined.

The time complexity of $n$-way sensitivity analysis may be fairly high for large $n$, even with the methods presented in this paper. Also, our method assumes that the variables and the parents associated with the $n$ parameters reside in the same clique in a junction tree. The method of Coupé et al. (2000) for $n$-way sensitivity analysis is based on propagation of tables of coefficients in a junction tree, and, therefore, has a (potentially very much) larger space requirement. However, their method is general in the sense that it does not put any restrictions on the location of the parameters.

**Acknowledgements** During the initial phase of the work, the authors received valuable comments from Finn V. Jensen. Also, he suggested the basis for the general method for $n$-way analysis, described at the end of Section 6.

The research has been partly funded by the Danish National Centre for IT research, Project no. 87.2.